# Demo of Sanskrit-Hindi SMT System

**Rajneesh Pandey, Atul Kr. Ojha, Girish Nath Jha**
Jawaharlal Nehru University
New Delhi, India
{rajneeshp1988, shashwatup9k, girishjha}@gmail.com

**Abstract**

The demo proposal presents a Phrase based Sanskrit-Hindi (SaHiT) Statistical Machine Translation system. The system has been developed on Moses. 43k sentences of Sanskrit-Hindi parallel corpus and 56k sentences of a monolingual corpus in the target language (Hindi) have been used. This system gives 57 BLEU score.

## 1. Introduction

Sanskrit and Hindi belong to an Indo-Aryan language family. Hindi is considered to be a direct descendant of an early form of Sanskrit, through Sauraseni Prakrit and 1 speaker in India. Today Hindi is widely spoken across the country as well as in some parts of countries like Mauritius etc. According to the Census of 2001[1], India has more than 378,000,000 Hindi speakers.

The knowledge or information source can be accessed by users through translation of the texts from Sanskrit to other languages. Development of a Machine Translation (MT) system like Sanskrit-Hindi (SaHiT) MT can provide faster and easy solution for this task. Therefore, it becomes necessary that the knowledge contained in Sanskrit texts should be translated in Hindi in easy and cost-effective ways. At Present, there are many online MT systems available for Indian languages like Google, Bing Anussaraka, Anglabharati etc. but not for Sanskrit-Hindi. Even lesser work has been done for building SaHiT MT system: (a) "Development of Sanskrit Computational Tools and Sanskrit-Hindi Machine Translation System (SHMT)"[2] project (from April 2008-March 2011), sponsored by Ministry of Information Technology, Government of India, Delhi. It was a consortia project and in the project, 10 institutions/universities were involved[3] and they followed rule-based approach. It is not fully functional and not accessible. And. (b) During the PhD research work, Pandey (2016) has developed SaHiT MT system on Microsoft Translator Hub (MTHub) and Moses platforms. The MT system achieved 35.5 BLEU score on MTHub but it is also not accessible yet. Details study of MT hub training; error analysis and evaluation were reported in Pandey et.al (2016) and Pandey (2016).

Hence, we have demonstrated in this demo only Phrase based Machine Translation (PBSMT) of SaHiT MT system which was trained on Moses.

## 2. Description of Moses based SaHiT MT System

The first step was the creation of parallel (Sanskrit-Hindi) corpus and monolingual corpus of the target language (Hindi). We prepared 43k sentences. Out of 43k, 25k sentences were collected from the Dept. of Public Relation[4], Madhya Pradesh (MP) government and rests of the data were manually translated. The next step was the collection of a monolingual corpus and 56k sentences were crawled. The detailed statistics are presented below (Table 1):

| Sources | Parallel (sentences) | Monolingual (sentences) |
|---|---|---|
| News domain | 25 K | 49 K |
| Literature domain (including Panchtantra stories, books & sudharma journal etc. ) | 18 K | 2 K |
| Health and Tourism domain | 0 | 5 K |
| **The total size of the corpus** | 43 K | 56 K |

Table 1 Statistics of Corpus

After collection of data, we conducted several experiments using Moses tool to get good results. The Moses is an open source SMT toolkit which gives permission to automatically train translation model for any language pair i.e. Sanskrit and Hindi Kohen et. al (2007).

For building the system, we followed the processes of tokenization of parallel and monolingual corpus, filtering out long sentences, the creation of language and translation model, tuning, testing, automatic and human evaluation.

Figure 1 demonstrates user interface of the SaHiT MT system. Initially, user gives input text or uploads Sanskrit text file. Once it has entered or uploaded, it goes for preprocessing such as identification of source language and tokenization. When it is finished, input text goes to tuned model where the model file generates target text output. After that translated sentences go for detokenization and that will be displayed on web interface.

---

[1] https://www.ethnologue.com/language/hin

[2] http://sanskrit.jnu.ac.in/projects/shmt.jsp

[3] http://sanskrit.jnu.ac.in/projects/SHMT_images/SHMT_PI.pdf

[4] http://mpinfo.org/News/SanskritNews.aspx

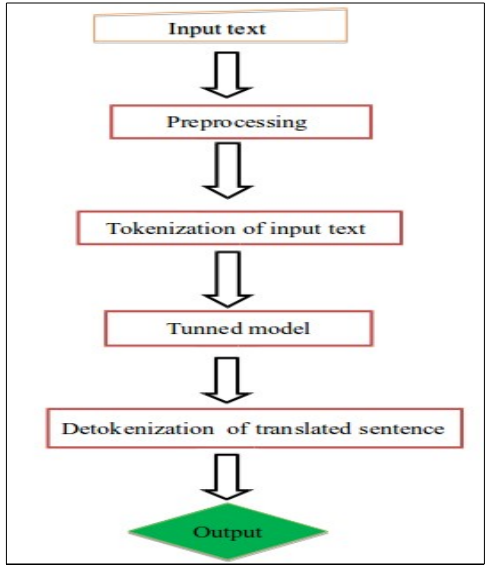

**Figure1: Architecture of SaHiT MT System**

## 3. Evaluation

The best MT system was developed after several experiments. Here, we present best of three experiments results of SaHiT MT system.

| Experiment phase | Parallel corpus | Monolingual corpus | BLEU score |
|---|---|---|---|
| First | 10K | 15K | 42 |
| Second | 26K | 40K | 54 |
| Third | 43K | 56K | 57 |

**Table 2: Automatic evaluation of SaHiT MT in various phases**

In third phase experiments, we have got 57 BLEU score. We have also evaluated on human evaluation parameter to the last phase experiment. This Sanskrit-Hindi MT system was evaluated by three evaluators. They judged the MT output based on the adequacy and fluency. Adequacy and fluency are calculated based on score between 1-5 given by the evaluators. 91% Adequacy and 66.72% Fluency.

The system can be accessed at the following web link: http://sanskrit.jnu.ac.in/index.jsp. Some examples of the MT output are presented below:

i. ते सन्तः वैश्या: सन्ति, ते कदापि न्यूनं न अमान् । (IS[5])

   वे सच्चे वैश्य हैं , उन्होंने कभी कम नहीं तोला । (MO)

---

[5] IS= Input Sentence, MO=MT Output

RT= Reference Translation

वे सच्चे वैश्य हैं , उन्होंने कभी कम नहीं तोला । (RT)

ii. मदमुक्तं समाजस्य सकंल्प: नयेयु:। (IS)

   नशामुक्त समाज का सकंल्प लें । (MO)

   नशामुक्त समाज का सकंल्प लें । (RT)

### 3.1 Error Analysis -

The MT system encountered several errors. But during the linguistics evaluation, we found the system is not able to produce correct output of target language in the case of Karka relational sentences, Complex sentences, and with Compounding and Sandhi words which reduced the systems accuracy around 68.43 out 100% Pandey 2016). It happens because the system was trained on very small size of corpus. So far this reason, the system is not able to generate. For example:

**(a) Issues in Karka level**

प्रदेशे महिला सशक्तिकरणाय योजना: चालयन्ते । (IS)

प्रदेशे में महिला सशक्तिकरणाय योजना चालये जा रहे हैं । (MO)

प्रदेशे में महिला सशक्तिकरण के लिये योजनाएँ चालयी जा रही हैं । (RT)

**(b) Issues with complex sentences**

महिला–बाल विकास मन्त्रिणी श्रीमती माया सिंह ग्वालियरे नारी निकेतनत: एकादशमहिलाया पलायिता घटनां गंभीरताया नेयितुम् आयुक्ता मिहला सशिक्तकरणं निरीक्षणं करणस्य निर्देश: दत्तवन्त: सन्ति । (IS)

महिला–बाल विकास मन्त्रिणी श्रीमती माया सिंह ग्वालियर नारी निकेतनत के एकादशमहिलाया भाग गयी घटना का स्मरण गंभीरता से लें आयुक्त महिला सशिक्तकरण निरीक्षण करने के  निर्देश दिये गये हैं । (MO)

महिला–बाल विकास मन्त्री श्रीमती माया सिंह ने ग्वालियर में नारी निकेतन से 11 महिला के भाग जाने की घटना को गंभीरता से लेते हुए आयुक्त मिहला सशिक्तकरण को जाँच करने के  निर्देश दिये हैं । (RT)

## 4. Conclusion and Future work

SaHiT attempts to translate Sanskrit text into Hindi language. It gives decent results as compared to previous rule-based MT system or others. Now days it produces 91% adequacy and 66.72% fluency. In future, we will collect more data to train on NMT approach and also work on improving the translation quality of complex and long sentences, and compounding problems etc.